\pdfoutput=1

\documentclass[11pt]{article}

\usepackage{acl}

\usepackage{times}
\usepackage{latexsym}

\usepackage[T1]{fontenc}

\usepackage[utf8]{inputenc}

\usepackage{microtype}

\usepackage{inconsolata}

\usepackage{graphicx}

\usepackage{booktabs}
\usepackage{fixltx2e} 
\usepackage{ulem}
\renewcommand{\emph}[1]{\textit{#1}}
\usepackage{subcaption}
\usepackage{amsmath}
\usepackage{amsfonts}
\usepackage{multirow}
\usepackage{bbm}
\usepackage{algorithm}
\usepackage{algpseudocode}

\usepackage[utf8]{inputenc}
\usepackage{tcolorbox}
\usepackage{authblk}

\title{GeoFM: Enhancing Geometric Reasoning of MLLMs via Synthetic Data Generation through Formal Language}

\author{
\text{Yuhao Zhang}, \text{Dingxin Hu}, \text{Tinghao Yu}, \text{Hao Liu}, \text{Yiting Liu} \\
Tencent Hunyuan Team \\
\texttt{\{yuhaozzhang,dingxinhu,maxwellyu,paulhliu,timytliu\}@tencent.com} \\
}

\begin{document}
\maketitle
\begin{abstract}
Multi-modal Large Language Models (MLLMs) have gained significant attention in both academia and industry for their capabilities in handling multi-modal tasks. However, these models face challenges in mathematical geometric reasoning due to the scarcity of high-quality geometric data. To address this issue, synthetic geometric data has become an essential strategy. Current methods for generating synthetic geometric data involve rephrasing or expanding existing problems and utilizing predefined rules and templates to create geometric images and problems. However, these approaches often produce data that lacks diversity or is prone to noise. Additionally, the geometric images synthesized by existing methods tend to exhibit limited variation and deviate significantly from authentic geometric diagrams. To overcome these limitations, we propose GeoFM, a novel method for synthesizing geometric data. GeoFM uses formal languages to explore combinations of conditions within metric space, generating high-fidelity geometric problems that differ from the originals while ensuring correctness through a symbolic engine. Experimental results show that our synthetic data significantly outperforms existing methods. The model trained with our data surpass the proprietary GPT-4o model by 18.7\% on geometry problem-solving tasks in MathVista and by 16.5\% on GeoQA. Additionally, it exceeds the performance of a leading open-source model by 5.7\% on MathVista and by 2.7\% on GeoQA.
\end{abstract}

\section{Introduction}

\begin{figure}[t]
    \centering
    \includegraphics[scale=0.28]{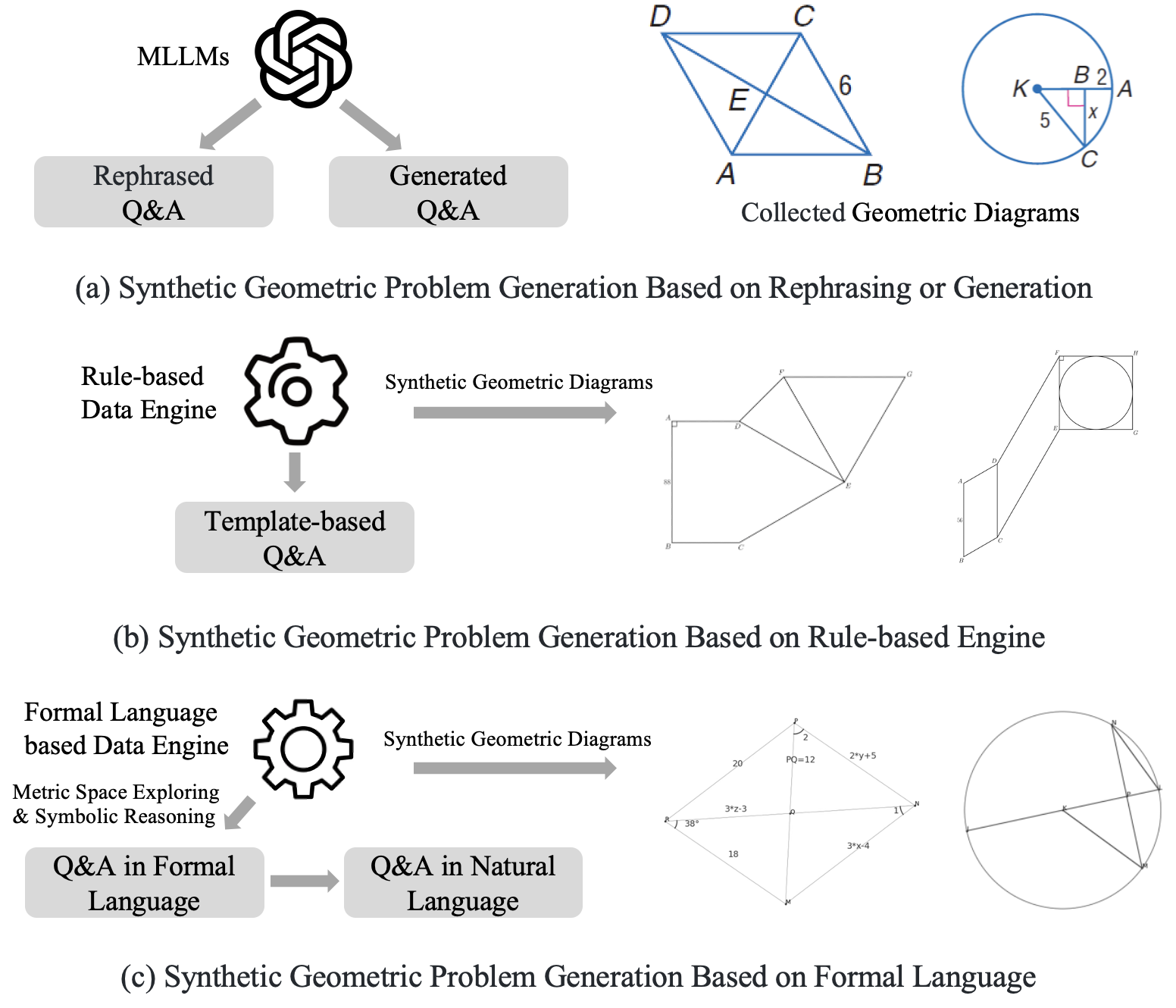}
    \caption{ 
    Comparison of different methods for synthesizing geometric data.
    (a) Generate geometric Q\&A data by using MLLMs to rephrase existing problems or create new Q\&A from collected geometric images.
    (b) Utilize a rule-based data engine to generate template-based Q\&A and low-fidelity images.
    (c) Employ formal language to explore the combinations of geometric metric conditions and synthesize new problems, ensuring solution accuracy through symbolic reasoning, and generate high-fidelity geometric images.
    }
    \label{fig:intro}
\end{figure}

Large language models (LLMs) exhibit excellent reasoning capabilities. There has been a significant amount of research dedicated to applying large language models to solve text-based mathematical problems, resulting in substantial progress \citep{openai2024gpt4ocard, luo2023wizardmath,shao2024deepseekmathpushinglimitsmathematical,yang2024qwen25mathtechnicalreportmathematical}. Recently, there has also been a growing focus on using Multi-modal Large Language Models (MLLMs) to address multi-modal mathematical problems that include images \citep{gao2023gllavasolvinggeometricproblem,shi-etal-2024-math,zhang2024mavismathematicalvisualinstruction,li2024eagleelevatinggeometricreasoning}. Although MLLMs perform well in general tasks such as Visual Question Answering (VQA), their performance often falls short when tackling multi-modal mathematical problems \citep{lu2024mathvistaevaluatingmathematicalreasoning,wang2024measuringmultimodalmathematicalreasoning}. In particular, geometry problems, which are a typical example of multi-modal mathematical problems with wide-ranging applications, require the integration of both visual and textual information for reasoning and solution. However, MLLMs struggle with these problems. One of the primary reasons for this difficulty is the lack of high-quality geometric data for training MLLMs. Compared to natural scene tasks like VQA, the sources and quantity of geometric data are relatively limited, which hinders the advancement of MLLMs' abilities in geometry.

To address the shortage of geometric data, some approaches have employed synthetic data generation. A straightforward method involves rewriting the problem statements and answers \citep{gao2023gllavasolvinggeometricproblem}. However, simple rewrites do not alter the underlying meaning of the problems. Although this increases the quantity of problems, it does not enhance the diversity. Other approaches have attempted to use MLLMs to modify original geometric problems and generate answers \citep{gao2023gllavasolvinggeometricproblem}, or to directly create new problems and corresponding responses based on collected geometric images \citep{shi-etal-2024-math}, as shown in Figure \ref{fig:intro}(a). Nevertheless, these methods rely on the geometric reasoning capabilities of MLLMs. Given the current limitations of MLLMs in solving geometric problems, these approaches are prone to introducing noise into the synthetic data. Recently, there have been attempts to synthesize geometric problems using predefined rules and templates \citep{kazemi2023geomversesystematicevaluationlarge, zhang2024mavismathematicalvisualinstruction}. For example, new shapes are generated by continuously extending basic geometric figures such as triangles and quadrilaterals outward along their edges. The reasoning paths and final answers are obtained through programming, as illustrated in Figure \ref{fig:intro}(b). While this method ensures the correctness of the reasoning and answers, the low fidelity of the synthesized images and the restricted variety of problems resulting in a significant disparity from real geometric problems. This discrepancy limits the progress of MLLMs in developing geometric capabilities.
 
To address the challenges present in current approaches, we propose a novel method for synthesizing geometric data. We have observed that existing geometric datasets often associate a single geometric diagram with only one or two problems, despite the fact that geometric diagrams often contain rich metric information that are not fully covered by the existing problems. Therefore, we propose GeoFM, a method that employs formal languages to explore the combinations of conditions within metric spaces of geometric diagrams, thereby generating high-fidelity geometric problems differ from the original ones but whose correctness is guaranteed using a symbolic engine. Existing work on geometric formal languages is scattered across different fields, such as geometric problem solving \citep{lu-etal-2021-inter, peng-etal-2023-geodrl, zhang2024formalgeoextensibleformalizedframework}, theorems proving \citep{Trinh2024SolvingOG, chervonyi2025goldmedalistperformancesolvingolympiad} and geometric drawing \citep{krueger2021automaticallybuildingdiagramsolympiad}. Furthermore, these studies frequently necessitate human intervention, such as manual formalization, to accomplish the associated tasks  \citep{zhang2024formalgeoextensibleformalizedframework, krueger2021automaticallybuildingdiagramsolympiad}, which prevents their application for large-scale automatic synthesis of geometric data. To address this issue, we propose a comprehensive framework for geometric data synthesis that automates the formalization of seed problems, the synthesis of new problems, and the generation of images. Utilizing this approach, we have developed a highly accurate and realistic geometric synthetic dataset GeoFM80K. Experimental results demonstrate our synthetic data can effectively enhance the geometric reasoning capabilities of MLLMs. 

Our contributions are summarized as follows:

1. We propose GeoFM, a geometric data synthesis method using formal languages and symbolic reasoning to generate accurate solutions and new geometric diagrams, addressing data noise and discrepancies in existing data synthesis methods.

2. We introduce a strategy for synthesizing new geometric problems through the combination of geometric metric conditions, resulting in the GeoFM80K dataset. Models trained on GeoFM80K outperform those trained on representative synthetic data by 8.2\% on MathVista-GPS \citep{lu2024mathvistaevaluatingmathematicalreasoning} and 11.1\% on GeoQA \citep{chen-etal-2021-geoqa}.

3. Experimental results show our method enhances the geometric reasoning of MLLMs. The GeoFM-8B model surpasses GPT-4o by 18.7\% on MathVista-GPS and 16.5\% on GeoQA, and exceeds the best open-source model by 5.7\% on MathVista-GPS and 2.7\% on GeoQA.

\begin{figure*}[t]
    \centering
    \includegraphics[scale=0.31]{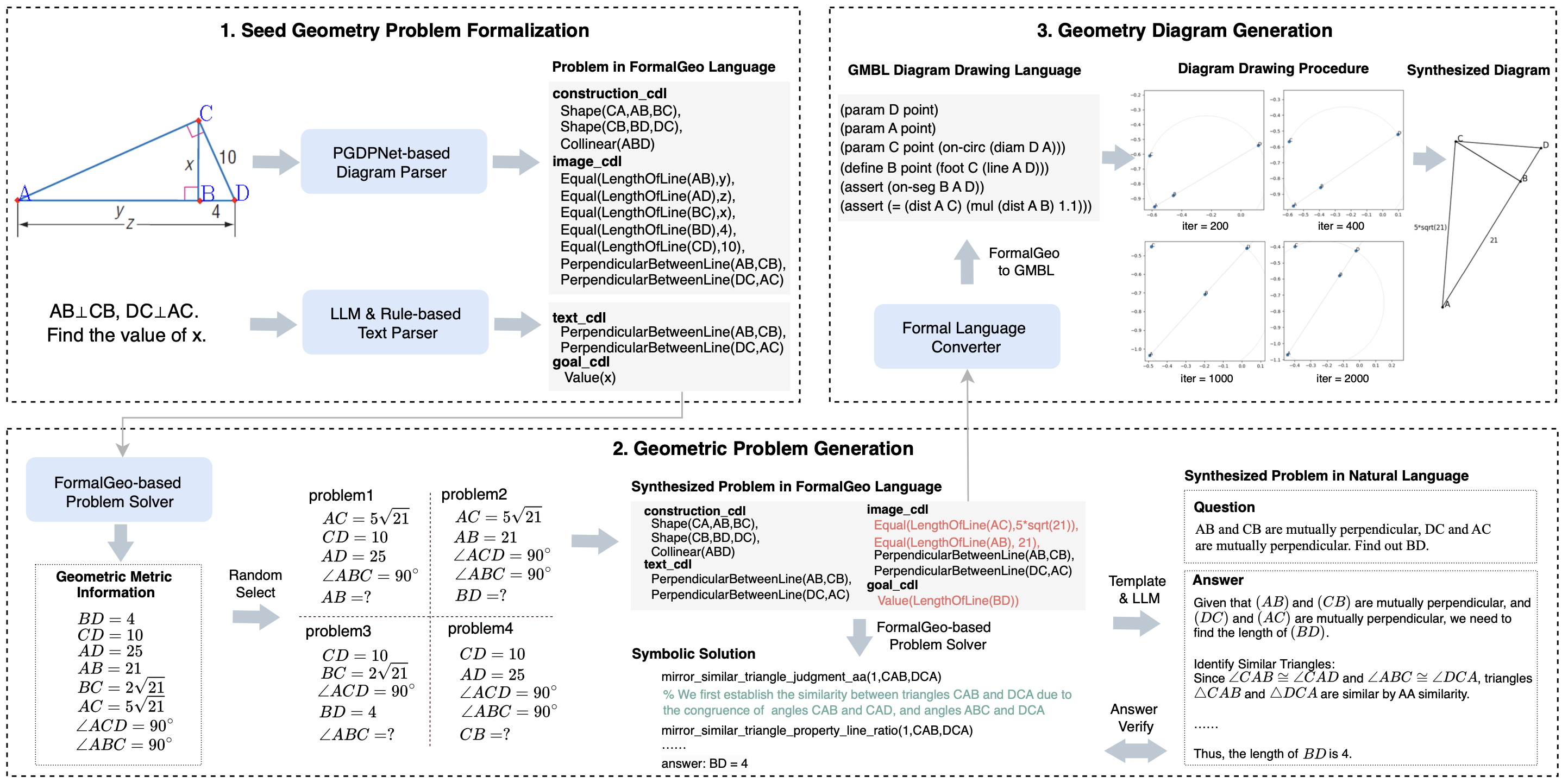}
    \caption{The Framework of Geometric Data Synthesis GeoFM}
    \label{fig:framework}
\end{figure*}

\section{Method}
\subsection{Overview}
In this section, we present our method for generating synthetic geometric problems. We start by converting seed problems into a formal language for problem-solving. New problems are created by combining metric conditions from the seed problems and solved using symbolic reasoning, enabling natural language solution synthesis and result verification. These formal representations are then translated into a drawing language to produce geometric diagrams. This process results in a synthetic dataset with solutions verified by a symbolic engine and newly synthesized diagrams, ensuring data accuracy and diversity. The framework is illustrated in Figure \ref{fig:framework}.

\subsection{Seed Geometry Problem Formalization} \label{formal}
Formalizing geometric problems is a significant research area in geometry. Various formalization schemes have been proposed, including InterGPS \citep{lu-etal-2021-inter}, AlphaGeometry \citep{Trinh2024SolvingOG, chervonyi2025goldmedalistperformancesolvingolympiad}, and FormalGeo \citep{zhang2024formalgeoextensibleformalizedframework}, each employing different approaches. In this study, we utilize FormalGeo as it more effectively represents metric geometry than AlphaGeometry and offers a broader range of geometric theorems than InterGPS. FormalGeo employs the Conditional Declaration Language (CDL) to represent geometric problems, which includes construction CDL, text CDL, image CDL, and goal CDL. Construction CDL conveys geometric structure information, such as basic shapes, collinearity, and cocircularity. Text CDL and image CDL capture geometric and algebraic relations from the problem statement and diagram, respectively, while goal CDL defines the problem-solving objective. An illustrative example is shown in Figure \ref{fig:framework}.

For the text parser, we propose a new construction method based on training a large language model with synthetic data. Since the text parser focuses on mapping natural language to formal language without considering the validity or solvability of the problem, we propose a method for generating synthetic training data based on formal language back-translation. Initially, for each formal language expression in FormalGeo, we use GPT-4o to generate 20 corresponding natural language templates, which are then manually reviewed and corrected. During data synthesis, we randomly select formal language conditions and goals to be solved, insert randomly generated geometric points to create a formal language problem, and then convert it into a natural language problem description using the natural language templates. This description is rewritten using the large language model Qwen2.5-72B-Instruct \citep{qwen2025qwen25technicalreport} to increase the diversity of expressions. In this way, we construct synthetic training data for the text parser that maps natural language problems to formal language problems. Using this method, we synthesized 30k training data samples and trained Llama-3-8B-Instruct \citep{grattafiori2024llama3herdmodels}, resulting in the development of a text parser.

For the diagram parser, we constructed it by integrating the geometric shape parsing method PGDPNet \citep{ijcai2022p0228}, OCR tool \citep{du2021ppocrv2bagtricksultra}, and rule-based processing. PGDPNet can identify various geometric elements, including points and lines, their coordinates, and geometric relationships like parallelism and perpendicularity. To enhance the accuracy of text and symbol recognition, we employ OCR to re-recognize the information within the detection boxes extracted by PGDPNet. Based on all the parsed information, we convert it into construction CDL and image CDL through rule-based processing.

The seed problems are processed using the text parser and the diagram parser to derive their formal representations. 
After filtering out invalid conditions using formal language grammar validation, seed problems represented in formal language are generated. These seed problems are then used for subsequent geometric problem synthesis. 
It is important to note that while parsing errors by the text parser and diagram parser may cause discrepancies between the formalized problems and the original ones, the final synthesized data remains consistent and error-free. This is because both the new problems and the corresponding images are generated solely based on the formalized seed problems, rather than the original ones.

\subsection{Geometric Problem Generation} \label{generation}
In this section, we will introduce the process of generating new geometry problems based on formalized seed problems. Since each geometric diagram contains rich metric information such as lengths, angles, and areas, we can utilize the formal language representation to combine the metric information in various ways, thereby generating new problems with different conditions and goals. Specifically, the synthesis process primarily consists of three components: calculating the geometric metric information of the seed problems, synthesizing data in formal language, and converting this data into natural language geometric instruction data. The process is detailed in Algorithm \ref{alg1}.

\subsubsection{Gathering Geometric Metrics} \label{cal}
To extract as much metric information as possible from the seed problems, we utilize the FormalGeo problem solving engine. During the solving process, we employ a breadth-first search approach to determine the applicability of predefined geometric theorems to the problems, continuing until a solution is found or a timeout occurs. Regardless of whether the solution is ultimately successful, the reasoning process yields substantial metric information about various geometric elements in the problem. We extract this metric information \( \mathcal{M}_{all} \) for the subsequent synthesis of new problems.

\captionsetup[algorithm]{font=small}
\begin{algorithm} 
\small
\caption{Geometric Problem Generation}

\label{alg1}
\begin{algorithmic}[1]
\renewcommand{\algorithmicrequire}{\textbf{Input}}
\renewcommand{\algorithmicensure}{\textbf{Output}}
\Require formalized seed problem set $\mathcal{FS}$, number of synthetic problems $m$
\Ensure synthetic problem set $\mathcal{S}$
\For{$\mathcal{P} \in \mathcal{FS}$}
    \State $\mathcal{M}_{p} \gets \text{MetricInfoOfProblemStatement}(\mathcal{P})$
    \State $\mathcal{M}_{all} \gets \text{GatheringMetricInfo}(\mathcal{P})$
    \State $m_p \gets m$
    \While{$m_p > 1$}
        \State $n \gets \text{Random}\bigl(1,\ \min\bigl(\lvert \mathcal{M}_p\rvert,\ \lvert \mathcal{M}_{all}\rvert - \lvert\mathcal{M}_p\rvert\bigr)\bigr)$
        \State $\mathcal{M}_{del} \gets \text{RandomSelect}(\mathcal{M}_p,\, n)$
        \State $\mathcal{M}_{add} \gets \text{RandomSelect}(\mathcal{M}_{all}\setminus \mathcal{M}_{p},\, n)$
        \State $\mathcal{P}_{new} \gets \mathcal{P} \setminus \mathcal{M}_{del} \cup \mathcal{M}_{add}$
        \State $\mathcal{A}_{new} \gets \text{FormalGeoSolver}(\mathcal{P}_{new})$
        \State $\mathcal{P}_{syn},\, \mathcal{A}_{syn} \gets \text{Template\&LLM}(\mathcal{P}_{new},\, \mathcal{A}_{new})$
        \If{$\text{AnswerVerify}(\mathcal{A}_{syn},\,\mathcal{A}_{new})$}
            \State $\mathcal{S}.\text{add}([\mathcal{P}_{syn},\,\mathcal{A}_{syn}])$
            \State $m_p \gets m_p - 1$
        \EndIf
    \EndWhile
\EndFor
\State \textbf{return} $\mathcal{S}$
\end{algorithmic}
\end{algorithm}

\subsubsection{Synthesizing Data in Formal Language}
After obtaining geometric metric conditions \( \mathcal{M}_{all} \) for a seed problem \( \mathcal{P} \), we can combine these conditions to generate new geometric problems. Let \( \mathcal{M}_{p} \) be the set of metric conditions of the original problem statement. We first sample a random number $n$ (where $ n \leq \min(|\mathcal{M}_p|, |\mathcal{M}_{all}| - |\mathcal{M}_p|) $). Next, we replace \( n \) metric conditions from \( \mathcal{M}_p \) with \( n \) new conditions sampled from the remaining metric set \( \mathcal{M}_{all} - \mathcal{M}_p \) and randomly choose one metric condition different from the new problem statement as the goal, thereby creating a new problem. This ensures that the new problem has the same number of metric conditions as the seed problem, minimizing issues related to insufficient metric conditions for deriving valid conclusions and avoiding redundancy from having too many conditions.
Furthermore, we randomly allocate the metric conditions to text CDL and image CDL. The metric conditions in image CDL will only appear in the synthesized images and not in the problem statements, thereby forcing the model to interpret the problem by reading the images rather than relying solely on textual information.

Once the formal language problem is obtained, we solve the synthesized problem using the FormalGeo symbolic engine to derive the corresponding symbolic solutions. The symbolic solution includes the geometric theorems applied and the derivation process. Since the goal of the synthesized problem is randomly selected and may not always be solvable, if the goal is not achieved, we select the last valid inference from the symbolic engine's reasoning path as the new goal. This ensures the validity of the problem. Through this process, we can synthesize multiple formal language problems with symbolic solutions from each seed problem.

\subsubsection{Geometric Instruction Data Synthesis}
After obtaining the formalized problems and their symbolic solutions, it is necessary to convert them into natural language instruction data to facilitate subsequent training of the MLLMs. This conversion process begins by transforming all FormalGeo formalized language and the geometric theorems used in problem-solving into natural language templates. These templates are manually verified to ensure their accuracy. Subsequently, we use these templates to convert the formalized problems and their symbolic solutions into natural language.

The lack of diversity in template-based solutions can lead to mode collapse when used directly for model training.
To address this issue, we employ the large language model Qwen2.5-72B-Instruct to rewrite the template-generated solutions, producing more fluent and varied problem-solving solutions. The prompt for rewriting is provided in Appendix \ref{prompt}. To minimize rewriting errors, we also use the LLM to compare the final answers of the rewritten problems with the results derived from FormalGeo through answer extraction and verification following the MathVista \citep{lu2024mathvistaevaluatingmathematicalreasoning} evaluation methodology, retaining only those problems where the answers are consistent. Compared to directly generating problem solutions using a strong MLLM, our method references the reasoning process of a symbolic engine during solution generation and the final answers are cross-verified for consistency with the results from the symbolic engine, thereby significantly reducing the probability of errors in the synthesized problem solutions. 

\subsection{Geometry Diagram Generation} \label{diagram}
Synthesizing geometric images for each generated problem is challenging due to the need to meet geometric constraints. Some methods use specialized drawing programs, but these often produce a limited variety of images that conform to predefined patterns \citep{kazemi2023geomversesystematicevaluationlarge, zhang2024mavismathematicalvisualinstruction}. Tools like GeoGebra \citep{hohenwarter2007} require manual manipulation for drawing. The Geometry Model Building Language (GMBL) \citep{10.1007/978-3-030-79876-5_33} uses a formal language and computational geometry to approximate target images through numerical optimization. However, it requires manually creating the formal language for the target image and evaluating if the synthesized image meets expectations, making it impractical for large-scale automated synthesis.

To address the limitations of existing methods, we developed a new engine capable of automatically synthesizing large-scale geometric images based on GMBL. This engine contains a formal language converter that automatically transforms construction CDL and image CDL statements, which illustrate geometric diagrams, into GMBL formal language. This conversion requires the prior construction of a mapping table from the FormalGeo language to the GMBL language. When generating the GMBL description of a problem, a heuristic rule-based method is first employed to determine the definition order of geometric points. Subsequently, the relevant geometric constraints represented in the FormalGeo language for each geometric point are translated into the GMBL language based on predefined rules and the mapping table.

We categorize the computational geometry objects in GMBL used to assess whether geometric constraints are met based on the strictness of these constraints. For example, the requirement for a point to lie on a line is stricter than that for two line segments to be of equal length, as deviations from the former are more apparent. We then establish different loss thresholds for each group, filtering out images that do not meet these thresholds after numerical optimization to maintain the quality of synthetic images. For geometric images that satisfy the constraints, we incorporate image CDL information, such as segment lengths and angles, into the diagram. This inclusion ensures that MLLMs must interpret the image to extract necessary information for problem-solving, thereby enhancing the model's image perception capabilities. This approach allows us to automatically generate images corresponding to synthesized geometric problems represented by the FormalGeo formal language.

\section{Experiments}
\subsection{Experimental Setup}
We synthesized 80k data points for our experiments based on the training sets of the FormalGeo7K \citep{zhang2024formalgeoextensibleformalizedframework} and PGPS9K \citep{10.24963/ijcai.2023/376} geometric datasets. Synthetic images are generated with a 4:3 aspect ratio, where the shorter edge is randomly chosen to be either 112, 224, or 336 pixels in length. The effectiveness of our synthesized data was validated using the LLaVA-NeXT-8B \citep{liu2024llavanext}, a model trained with limited geometric data, which facilitates the assessment of how the addition of various geometric data affects the model's geometric capabilities. Additionally, we employed InternVL2-8B-MPO \citep{wang2024enhancingreasoningabilitymultimodal}, a model trained with a larger amount of geometric data, to determine whether synthesized data can further enhance the performance of models with higher geometric capabilities. Both models were trained with full-parameter tuning for two epochs, with detailed hyper-parameters provided in Appendix \ref{ap:hyper}. We utilized two most widely adopted benchmarks for evaluation: the MathVista for geometry problem-solving (GPS) \citep{lu2024mathvistaevaluatingmathematicalreasoning} and the GeoQA \citep{chen-etal-2021-geoqa}. Model performance was assessed through response generation, answer extraction, and score calculation, following the MathVista methodology.
Top-1 accuracy was used as the evaluation metric.

\begin{table} 
    \small
    \centering
    \begin{tabular}{l|cc} 
    \toprule
    Model & $D_{origin}$ & $D_{syn}$ \\ 
    \midrule
    LLaVA-NeXT-8B  & 11.2 & 9.5  \\
    Qwen2-VL-7B  & 28.2 & 15.8  \\ %
    InternVL2-8B-MPO  & 40.7 & 27.7 \\ %
    Qwen2-VL-72B & 38.1 & 28.9 \\
    InternVL2-Llama3-76B &32.9 & 28.5 \\
    GPT-4o & 39.2 & 36.6 \\
    Gemini-2.0-Flash-Thinking-Exp &	\bf{57.8} & \bf{40.5} \\
    \bottomrule
    \end{tabular}
    \caption{Comparison of MLLM performance on open source geometric data $D_{origin}$ and synthetic data $D_{syn}$.}
    \label{tab:comparison}
\end{table}

\subsection{Necessity of Metric Space Exploration}
Some MLLMs are trained using open-source geometric datasets, where each image is associated with only a few questions. This raises the question of whether MLLMs can generalize to other variations of questions related to the same geometric diagram. To investigate this, we conducted an experiment using synthetic data. We sampled 500 questions each from two commonly used open-source geometric datasets, GeoQA \citep{chen-etal-2021-geoqa} and Geometry3K \citep{lu-etal-2021-inter}, to create a test set $D_{origin}$. Correspondingly, we generated a synthetic test set $D_{syn}$, by creating an equal number of problems based on $D_{origin}$ but with different conditions or problem-solving objectives.

As illustrated in Table \ref{tab:comparison}, all models, including both small and large open-source models in Qwen2-VL \citep{wang2024qwen2vlenhancingvisionlanguagemodels} and InternVL2 \citep{wang2024enhancingreasoningabilitymultimodal} series, as well as proprietary models like GPT-4o and Gemini-2.0-Flash-Thinking-Exp, demonstrated lower performance on synthetic data $D_{syn}$ compared to original data $D_{origin}$. The performance gap is quite significant, with three out of seven models showing a gap exceeding 10\%, the largest reaching 17.3\%. This indicates that many existing MLLMs struggle to generalize from known problems to related scenarios. The suboptimal performance on \(D_{syn}\), generated via metric space exploration, suggests that utilizing same large-scale data synthesis method in model training could enhance geometric capabilities. This hypothesis will be validated in subsequent sections.

\subsection{Effectiveness of GeoFM}

\begin{figure*}[h]
    \centering
    \includegraphics[scale=0.4]{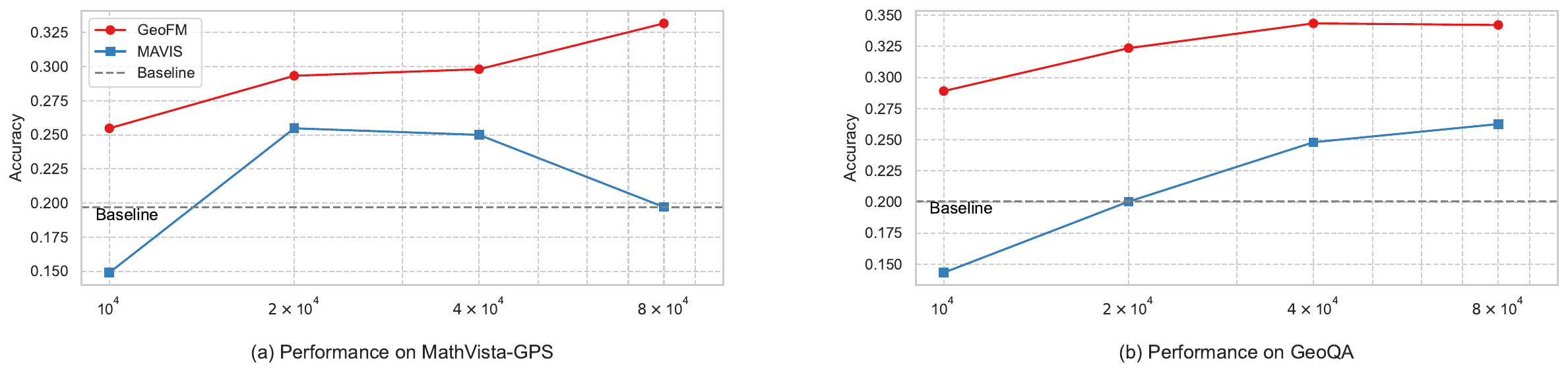}
    \caption{Comparison with existing geometric synthesis data at different data scales using LLaVA-NeXT-8B. The baseline corresponds to the performance of the original model.}
    \label{fig:scale}
\end{figure*}

\subsubsection{More Effective Utilization of Seed Data}
Effectively utilizing geometric seed data to enhance the geometric problem-solving abilities of MLLMs is a significant research question. In this section, we compare our GeoFM data synthesis method with various data construction approaches, including direct use of seed data, constructing chain of thought solutions based on GPT-4o \citep{openai2024gpt4ocard}, rewriting problems and CoT solutions, and augmenting problems and solutions with MLLMs as described by \citep{gao2023gllavasolvinggeometricproblem}. We sampled 5k geometric problems from the FormalGeo7K dataset as seed data and conducted experiments using LLaVA-NeXT-8B, training each dataset for two epochs as further training did not enhance performance. The results are presented in Table \ref{tab:seed-use}.

\begin{table} 
    \centering
    \footnotesize
    \begin{tabular}{l|l|cc} 
    \toprule
    \multirow{1}{*}{Training Data} & \multirow{1}{*}{Vol.} & \multicolumn{1}{|c}{MathVista} & \multicolumn{1}{c}{GeoQA} \\  %
    \midrule
    Base Model & & 19.7 & 20.0 \\
    \quad w/ Seed Data & 5k & 17.8 & 22.7  \\
    \quad w/ GPT-4o CoT & 5k & 25.9 & 22.9\\
    \quad w/ CoT + Rephrase  & 25k & 20.7 & 23.5	 \\
    \quad w/ CoT + MLLM Aug  & 25k & 26.3 & 25.8	 \\
    \quad w/ GeoFM Data & 25k & \bf{27.9} & \bf{32.0} \\
    \bottomrule
    \end{tabular}
    \caption{Results of different geometric seed data utilization methods on MathVista-GPS and GeoQA.}
    \label{tab:seed-use}
\end{table}

As demonstrated, utilizing GPT-4o's CoT data could enhance model performance. While simple rewrites show varying effectiveness across datasets, synthesizing new problems  improve performance.
The most significant improvement is achieved with the GeoFM data synthesis method, which increases performance by 10.1\% on the MathVista-GPS and 9.3\% on the GeoQA compared to the seed data. This indicates that our data synthesis method can more effectively utilize existing geometric data to help enhance model performance.

\subsubsection{Comparison with Existing Geometric Synthetic Datasets}
To assess the impact of using solely synthesized data, we compare GeoFM with existing geometric synthetic datasets. The GeoGPT4V \citep{cai-etal-2024-geogpt4v} dataset contains 4.9k synthetic data points, which is small in quantity. The GermVerse \citep{kazemi2023geomversesystematicevaluationlarge} dataset performs suboptimally on benchmarks. Therefore, our primary comparison is between GeoFM and the recently proposed MAVIS-Geometry \citep{zhang2024mavismathematicalvisualinstruction} dataset, a representative dataset generated through rule-based data engine. To evaluate the model's performance across various data scales, we sampled 10k, 20k, 40k, and 80k data points from each dataset. The experimental results presented in Figure \ref{fig:scale} evident that both datasets show performance improvements after training. However, GeoFM significantly outperforms MAVIS-Geometry, with an average improvement of 8.2\% on MathVista-GPS and 11.1\% on GeoQA. We speculate that this is primarily due to the rule-based  synthetic geometric problems in MAVIS-Geometry differing substantially from real data, as illustrated in Appendix \ref{app4}, thereby limiting its effectiveness.

\subsubsection{Performance Boost from GeoFM}
To assess the benefits of adding GeoFM synthetic data to existing open-source datasets, we conducted experiments using the Geo170K-QA \citep{gao2023gllavasolvinggeometricproblem} and MathV360K-GPS \citep{shi-etal-2024-math} geometric datasets. We trained two base models, LLaVA-NeXT-8B and InternVL2-8B-MPO, using both the open-source data alone and the open-source data combined with GeoFM data. The experimental results, presented in Table \ref{tab:improvement}, demonstrate that  models trained with the addition of GeoFM data achieved consistent improvements on the MathVista-GPS and GeoQA benchmarks. Specifically, LLaVA-NeXT-8B showed improvements of 1.9\% and 2.3\%, while InternVL2-8B-MPO exhibited gains of 4.8\% and 3.2\%, respectively.

\begin{table}[!t]
    \small
    \centering
    \begin{tabular}{l|cc} 
    \toprule
    Model & MathVista & GeoQA \\ 
    \midrule
    GM-LLaVA-NeXT-8B  & 54.8 & 68.3  \\
    GeoFM-LLaVA-NeXT-8B  & \bf{56.7} & \bf{70.6}  \\
    \midrule
    GM-InternVL2-8B-MPO  & 74.5 & 74.7 \\ 
    GeoFM-InternVL2-8B-MPO  & \bf{79.3} & \bf{77.9} \\
    \bottomrule
    \end{tabular}
    \caption{Performance Improvements from GeoFM: "GM-" models are trained on Geo170K-QA and MathV360K-GPS datasets; "GeoFM-" models incorporate an additional 80k GeoFM data.}
    \label{tab:improvement}
\end{table}

\begin{table}[!t] 
    \scriptsize
    \centering
    \begin{tabular}{lcc}
        \toprule
        \textbf{Model} & \textbf{MathVista} & \textbf{GeoQA} \\
        \midrule
        \multicolumn{3}{c}{Closed-source MLLMs} \\
        \midrule
        GPT-4o \citep{openai2024gpt4ocard} & 60.6 & 61.4 \\
        GPT-4V \citep{gpt4v} & 50.5 & - \\
        Gemini 1.0 Ultra \citep{geminiteam2024geminifamilyhighlycapable} & 56.2 & - \\
        \midrule
        \multicolumn{3}{c}{Open-source MLLMs} \\
        \midrule
        LLaVA-LLaMA-2-13B \citep{NEURIPS2023_6dcf277e} & 29.3 & 20.3 \\
        Qwen-VL-Chat-7B \citep{bai2023qwenvlversatilevisionlanguagemodel} & 35.6 & 26.1 \\
        InternVL2-Pro \citep{InternVL2} & 65.4 & - \\ 
        InternVL2-8B-MPO \citep{wang2024enhancingreasoningabilitymultimodal} & \underline{73.6} & 53.1 \\
        \midrule
        \multicolumn{3}{c}{Mathematical MLLMs} \\
        \midrule
        Math-LLaVA-13B \citep{shi-etal-2024-math} & 57.7 & 47.8 \\
        G-LLaVA-7B \citep{gao2023gllavasolvinggeometricproblem} & 53.4 & 62.8 \\
        MAVIS-7B \citep{zhang2024mavismathematicalvisualinstruction} & - & 66.7 \\
        EAGLE \citep{li2024eagleelevatinggeometricreasoning} & 54.3 & 67.1 \\
        GeoGPT4V \citep{cai-etal-2024-geogpt4v} & 64.4 & - \\
        GOLD \citep{zhang-moshfeghi-2024-gold} & - & \underline{75.2} \\
        \midrule
        GeoFM-8B & \bf{79.3} & \bf{77.9} \\
        \bottomrule
    \end{tabular}
    \caption{Comparison of GeoFM model with existing MLLMs on MathVista-GPS and GeoQA}
    \label{tab:performance}
\end{table}

\begin{figure*}[!t]
    \centering
    \includegraphics[scale=0.31]{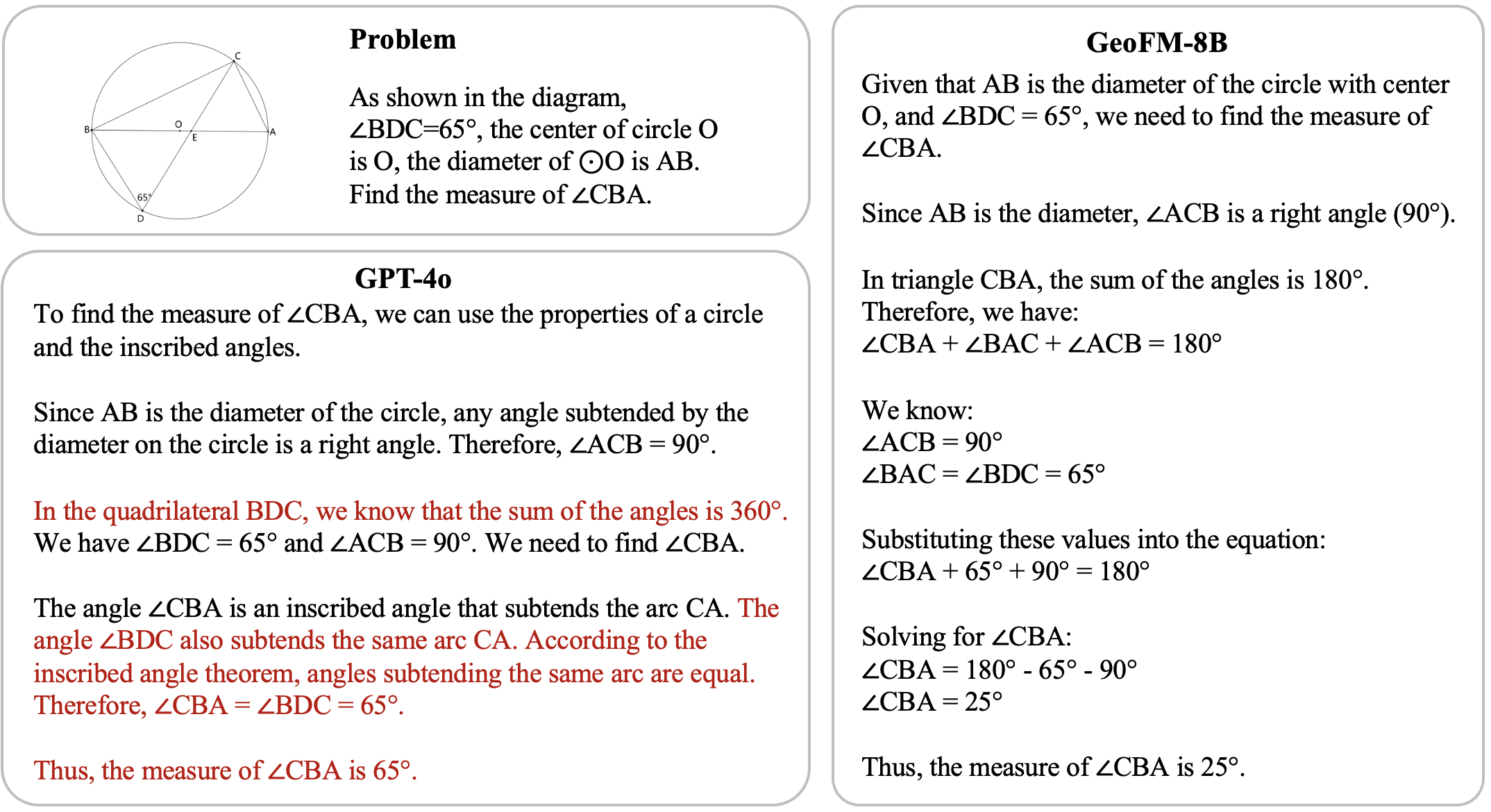}
    \caption{Demonstration of geometric problem solving using GPT-4o and GeoFM-8B}
    \label{fig:problem}
\end{figure*}

We compare GeoFM-8B which trained on the InternVL2-8B-MPO backbone with GeoFM data against existing MLLMs, including both proprietary and open-source representative models. The results, presented in Table \ref{tab:performance}, indicate that the GeoFM-8B model significantly outperforms existing models on the MathVista-GPS and GeoQA benchmarks. Specifically, it exceeds GPT-4o's accuracy by 18.7\% on MathVista-GPS and 16.5\% on GeoQA, and surpasses the leading model by 5.7\% on MathVista-GPS and 2.7\% on GeoQA.

We further validated our model's out-of-distribution (OOD) capabilities using the We-Math benchmark \citep{qiao2024wemathdoeslargemultimodal}. The experimental results indicate that our approach effectively generalizes to OOD dataset. See Appendix \ref{ood} for detailed results.

\subsection{Qualitative Analysis}
We conduct a qualitative analysis by comparing our model, GeoFM, with the representative model GPT-4o, as illustrated in Figure \ref{fig:problem}. Our model effectively captures the geometric features of the problems and provides an accurate reasoning process. In contrast, GPT-4o demonstrates errors in understanding geometric figures and  exhibits hallucinations that lead to incorrect answers. This comparison highlights the advantages of our synthetic  data method.

\section{Related Work}
\textbf{Geometry Problem Solving} Solving geometry problems is a challenging multi-modal mathematical task. Some studies have employed symbolic solvers to address geometric problems by first formalizing them and then performing symbolic reasoning \citep{lu-etal-2021-inter, li-etal-2024-lans,zhang2024formalgeoextensibleformalizedframework}. However, these symbolic solvers are limited to solving specific geometric problems and cannot transfer geometric capabilities across different scenarios like MLLMs. 
Recently, research aimed at enhancing the geometric capabilities of MLLMs has emerged, primarily by improving model performance through high-quality geometric data. Early geometric datasets such as GeoQA \citep{chen-etal-2021-geoqa}, GeoQA+ \citep{cao-xiao-2022-augmented}, UniGeo \citep{chen-etal-2022-unigeo}, and PGPS9K \citep{10.24963/ijcai.2023/376} were manually collected and curated, which often limited their scale. G-LLaVA \citep{gao2023gllavasolvinggeometricproblem} expanded existing geometric datasets using a large language model for rewriting and augmentation, but this method lacked diversity and was prone to introducing noise due to the limitations of the rewriting model. GeoGPT4V \citep{cai-etal-2024-geogpt4v} enhances this approach by incorporating image synthesis, generating Wolfram code via GPT-4 \citep{openai2024gpt4technicalreport}, and using this tool to create geometric images. However, this method's image synthesis is insufficiently stable.
GeomVerse \citep{kazemi2023geomversesystematicevaluationlarge} and MAVIS \citep{zhang2024mavismathematicalvisualinstruction} utilized rule-based data engines to generate geometric problems, but the data produced often differed significantly from real-world data, affecting their effectiveness. 
To address these shortcomings, we propose GeoFM, which employs formal languages to explore combinations of conditions within metric spaces, thereby generating high quality geometric data that can effectively enhance the geometric reasoning capabilities of MLLMs.

\section{Conclusion}
In this paper, we present GeoFM, a novel method for generating high-quality geometric problems to enhance the geometric reasoning abilities of MLLMs. GeoFM uses formal languages to systematically explore condition combinations within metric spaces. Our approach involves formalizing seed problems, generating new geometric problems through the combination of metric conditions, and creating geometric diagrams corresponding to the problems. Experimental results show that our method significantly outperforms existing approaches, achieving superior results on the MathVista and GeoQA benchmarks.

\section{Limitations}
In this study, we employ formal languages to explore various condition combinations within metric spaces of seed problems and synthesize high-quality geometric data to enhance the performance of multimodal large language models. During the synthesis process, we use seed problems to generate synthetic data, which need manual collection.  Additionally, certain types of geometric problems, such as word problems or those lacking geometric point identifiers, are challenging to formalize. Therefore, designing new methods for synthesizing geometric problems from scratch is a direction worth further exploration.

\bibliography{anthology,custom}
\appendix

\section{Hyper-parameters}  \label{ap:hyper}
The detailed hyper-parameters used for training LLaVA-NeXT-8B and InternVL2-8B-MPO are listed in Table \ref{tab:hyper}. We primarily adjusted the learning rate and batch size, while keeping the other parameters consistent with the original model's training configuration. All experiments are conducted using the Nvidia H20 graphics card, which has 96 GB of memory.

\begin{table} [th] 
    \footnotesize
    \centering
    \begin{tabular}{ll} 
    \toprule
    \textbf{Hyper-parameter} & \textbf{Value} \\ 
    \midrule
    \textbf{LLaVA-NeXT-8B} &  \\
    \quad training method & full parameter tuning \\
    \quad epochs & 2 \\
    \quad batch size & 64 \\
    \quad llm learing rate & 3e-5 \\
    \quad adapter learing rate & 3e-5 \\
    \quad vision tower learing rate & 2e-6 \\
    \quad vision select layer & -2 \\
    \quad warmup ratio & 0.03 \\
    \quad lr scheduler type & cosine \\
    \quad weight decay & 0 \\
    \midrule
    \textbf{InternVL2-8B-MPO} & \\
    \quad training method & full parameter tuning \\
    \quad epochs & 2 \\
    \quad batch size & 128 \\
    \quad llm learing rate & 1e-5 \\
    \quad adapter learing rate & 0 \\
    \quad vision tower learing rate & 0 \\
    \quad vision select layer & -1 \\
    \quad warmup ratio & 0.03 \\
    \quad lr scheduler type & cosine \\
    \quad weight decay & 0.01 \\
    \bottomrule
    \end{tabular}
    \caption{Hyper-parameters for model training}
    \label{tab:hyper}
\end{table}

\section{Performance on OOD Benchmark} \label{ood}
To assess the out-of-distribution (OOD) capabilities of our model, we utilized the newly introduced benchmark, We-Math \citep{qiao2024wemathdoeslargemultimodal}. 
This benchmark consists of manually curated data, independently collected and annotated according to a predefined knowledge structure, specifically designed to evaluate the reasoning abilities of MLLMs.
Our evaluation targeted plane geometry, including the "Calculation of Plane Figures" and "Understanding of Plane Figures" subfields. The experimental results, detailed in Table \ref{tab:ood}, indicate that our model demonstrated superior performance compared to recently proposed representative MLLMs. These findings suggest that our approach also possesses 
strong generalization capabilities on OOD dataset.
\begin{table}[ht]
\small
\centering
\begin{tabular}{lcc}
\toprule
\textbf{Model} & \textbf{CPF} & \textbf{UPF} \\
\midrule
G-LLaVA-13B \citep{gao2023gllavasolvinggeometricproblem} & 32.0 & 37.9 \\
Qwen-VL-Max \citep{Qwen-VL} & 39.8 & 41.4 \\
MiniCPM-LLaMA3-V2.5 \citep{yao2024minicpmv} & 40.8 & 39.8 \\
LLaVA-NeXT-72B \citep{liu2024llavanext} & 43.3 & 42.4 \\
InternVL2-8B-MPO \citep{wang2024enhancingreasoningabilitymultimodal} & 47.5 & 41.8 \\
GLM-4V-9B \citep{glm2024chatglm} & 51.3 & 46.5 \\
GeoFM-8B & \bf{52.2} & \bf{52.1} \\
\bottomrule
\end{tabular}
\caption{Comparison of the GeoFM model with existing MLLMs on the We-Math Benchmark. "CPF" indicates the "Calculation of Plane Figures" subfield while "UPF" indicates "Understanding of Plane Figures" subfield.} \label{tab:ood}
\end{table}

\clearpage

\onecolumn

\section{Template-based Solution Rewriting Prompt} \label{prompt}
\begin{tcolorbox}[colback=white,colframe=black!75!black,title=Prompt: Rewrite Template-based Solution]
Given a geometry problem and its answer hint, write a answer to the problem. Ensure the answer is correct, concise, easy to understand, and written with clarity and natural flow. \\

\textbf{Guidelines} \\
1. Refer to the answer hint, but do not use the information in it as given conditions. \\
2. Only output the solution, without any additional information. \\

\textbf{Problem} \\
<problem>\\

\textbf{Hint}\\
<template-based solution>

\end{tcolorbox} 

\bigskip

\section{Illustration of Geometric Problem and Solution Synthesis}
\begin{figure*}[!h]
    \centering
    \includegraphics[scale=0.47]{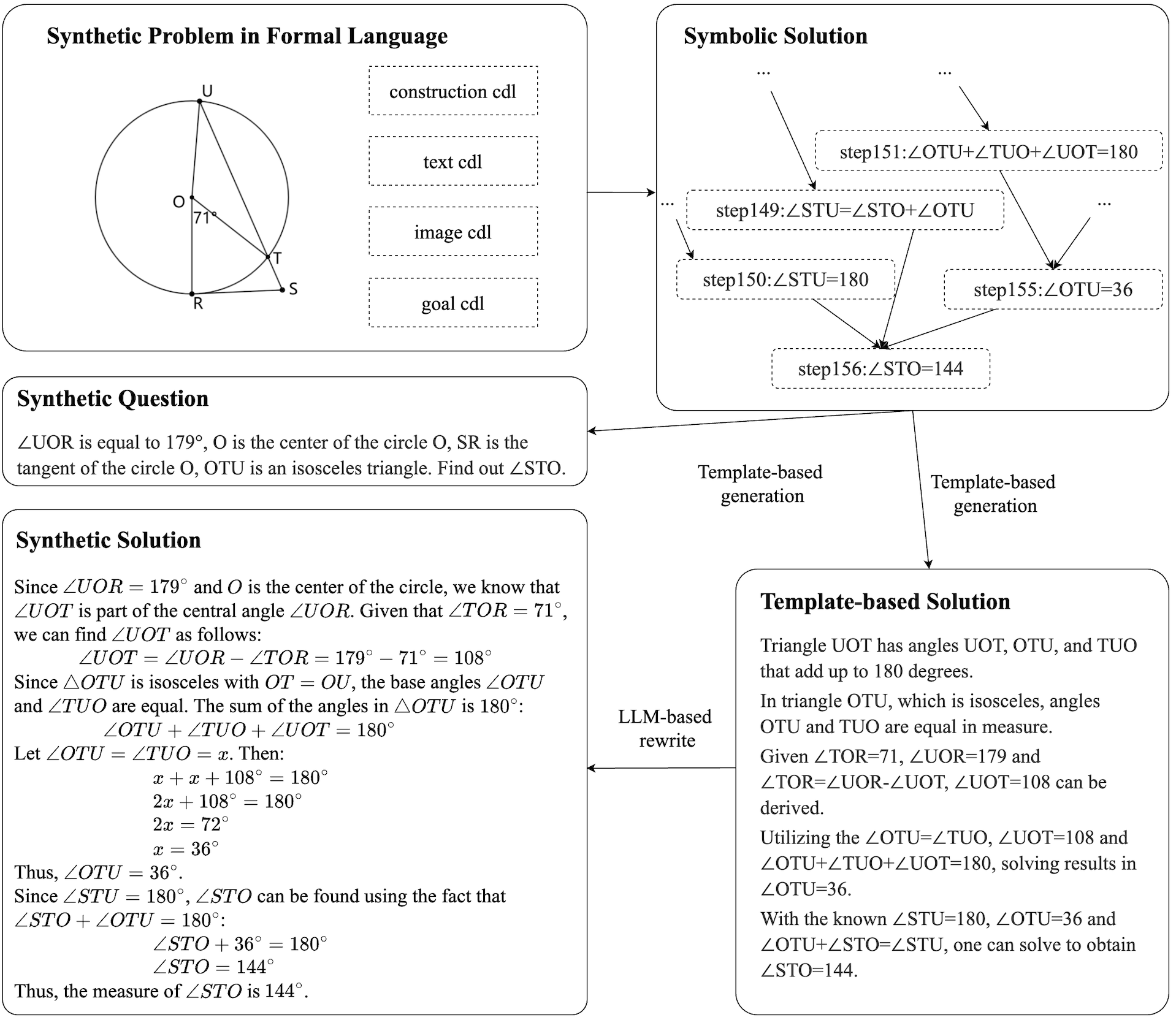}
    \caption{Convert a synthesized formal language geometric problem into natural language instruction data} 
    \label{fig:illustration}
\end{figure*}

\section{Examples of Synthetic Data}
\begin{figure*}[ht]
    \centering
    \includegraphics[scale=0.29]{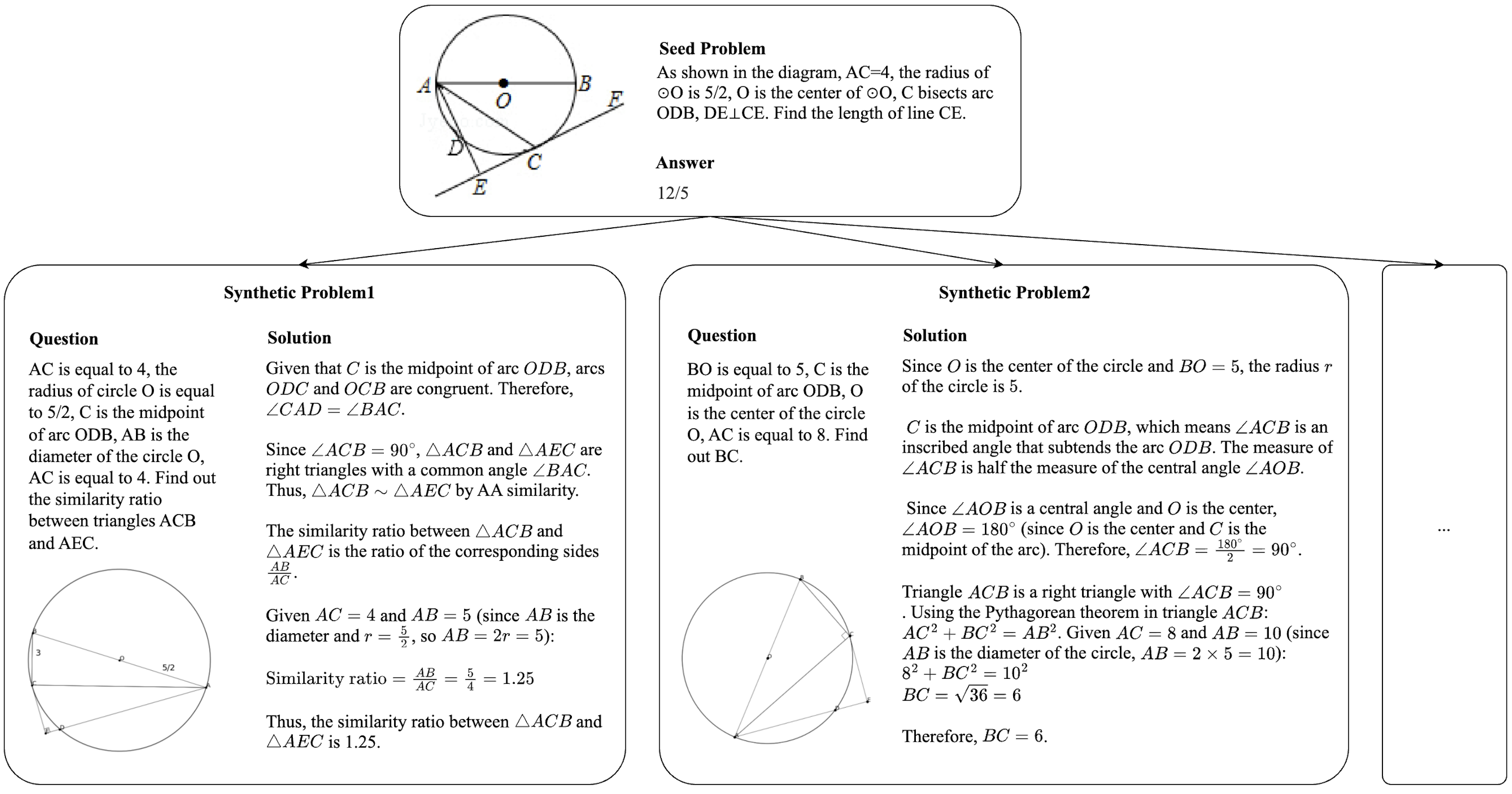}
    \caption{Examples of GeoFM Synthetic Data} 
    \label{fig:syn}
\end{figure*}

\section{Comparison of Geometric Images in Synthetic Datasets} \label{app4}
\begin{figure*}[ht]
    \centering
    \includegraphics[scale=0.43]{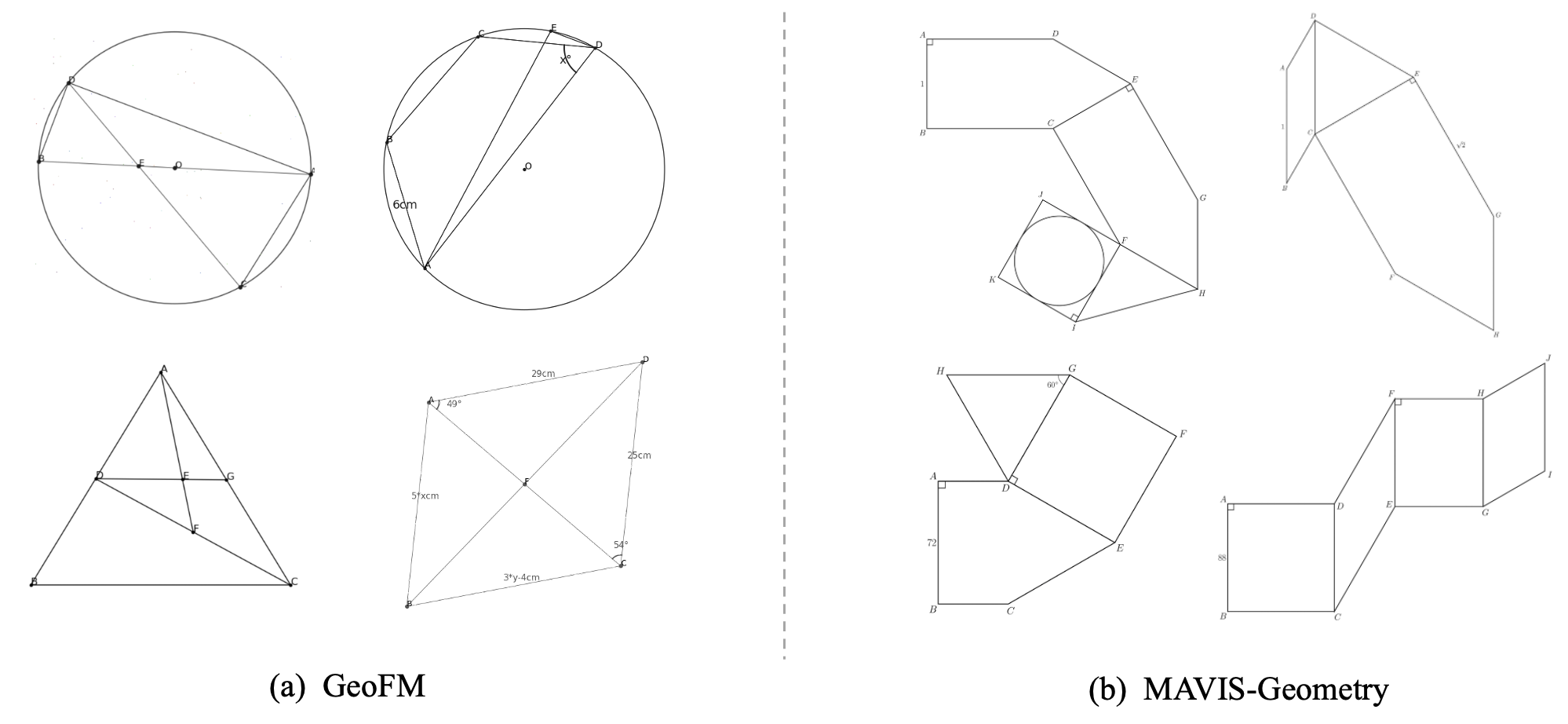}
    \caption{Comparison of Synthetic Images between GeoFM and MAVIS-Geometry} 
    \label{fig:comparison}
\end{figure*}

\end{document}